\documentclass[12pt]{elsarticle}
\usepackage{geometry}
\usepackage{lineno,hyperref}
\usepackage{amsmath}
\usepackage{amsfonts}
\usepackage[T1]{fontenc}

\usepackage{amssymb,amsthm,url,array}
\usepackage{algorithm,algcompatible}
\usepackage{float}
\usepackage{subfigure}
\usepackage{algpseudocode}
\usepackage{algorithmicx}
\usepackage{changes} 
\usepackage{tikz} 
\usepackage{xcolor}
\usepackage{dsfont}
\usepackage{verbatim}
\usepackage{color}
\usepackage{url}
\usepackage{multirow}
\usepackage{xurl}
\usepackage{soul}
\usepackage{booktabs}
\usepackage[title]{appendix}
\theoremstyle{plain}

\usepackage{graphicx}
\usepackage{caption}
\theoremstyle{definition}

\usepackage{arydshln} 
\usepackage{multirow}
\usepackage{bm}
\theoremstyle{remark}

\modulolinenumbers[1]
\usepackage{setspace}
\setcitestyle{authoryear,round}

\makeatletter
\def\ps@pprintTitle{%
    \let\@oddhead\@empty
    \let\@evenhead\@empty
    }
\makeatother
\journal{X}

\begin{document}
\begin{doublespace}
\begin{frontmatter}

\title{MoE-TransMov: A Transformer-based Model for Next POI Prediction in Familiar \& Unfamiliar Movements}
\author[1]{Ruichen Tan}
\author[1]{Jiawei Xue}
\author[2]{Kota Tsubouchi}
\author[3]{Takahiro Yabe}
\author[1]{Satish V. Ukkusuri\corref{cor1}}

\address[1]{Lyles School of Civil and Construction Engineering, Purdue University, West Lafayette, IN, USA}
\address[2]{LY Corporation, Tokyo, Japan}
\address[3]{Tandon School of Engineering, New York University, New York, NY, USA}

\cortext[cor1]{Corresponding author: sukkusur@purdue.edu}

\begin{abstract} 
Accurate prediction of the next point of interest (POI) within human mobility trajectories is essential for location-based services, as it enables more timely and personalized recommendations. In particular, with the rise of these approaches, studies have shown that users exhibit different POI choices in their familiar and unfamiliar areas, highlighting the importance of incorporating user familiarity into predictive models. However, existing methods often fail to distinguish between the movements of users in familiar and unfamiliar regions. To address this, we propose MoE-TransMov, a Transformer-based model with a Transformer model with a Mixture-of-Experts (MoE) architecture designed to use one framework to capture distinct mobility patterns across different moving contexts without requiring separate training for certain data. Using user-check-in data, we classify movements into familiar and unfamiliar categories and develop a specialized expert network to improve prediction accuracy. Our approach integrates self-attention mechanisms and adaptive gating networks to dynamically select the most relevant expert models for different mobility contexts. Experiments on two real-world datasets, including the widely used but small open-source Foursquare NYC dataset and the large-scale Kyoto dataset collected with LY Corporation (Yahoo Japan Corporation), show that MoE-TransMov outperforms state-of-the-art baselines with notable improvements in Top-1, Top-5, Top-10 accuracy, and mean reciprocal rank (MRR). Given the results, we find that by using this approach, we can efficiently improve mobility predictions under different moving contexts, thereby enhancing the personalization of recommendation systems and advancing various urban applications.

\end{abstract}
\end{frontmatter}

\section{INTRODUCTION}

With the rapid development of intelligent transportation systems and the widespread use of mobile devices, vast amounts of location-aware data are continuously generated through GPS tracking, mobile applications, and location-based services \citep{xu2022understanding}. Such data offer unprecedented opportunities for analyzing human mobility patterns, which are fundamental for transportation planning, urban management, and personalized travel services \citep{wang2019discovery}, which also offer a valuable lens for understanding how individuals interact with complex and heterogeneous urban environments, revealing spatial preferences, activity participation, and region-specific movement behaviors \citep{batty2013new, liu2014spatial}. Meanwhile, with the improvement of people's living standards and the increasing demand for personalized services, data-driven services have become increasingly important in providing accurate and personalized recommendations, such as user point of interest (POI) recommendations, targeted advertising, and public transportation recommendation \citep{albino2015smart, wang2023urban, chen2013travel}. In this case, POI recommendations is viewed as an effective analytical framework for modeling individual movement decisions and activity participation across different urban regions \citep{chen2016urban}.To provide more precise and personalized recommendations in POI systems, it is crucial to better understand human movement behaviors. Mobile check-in data and digital footprints provide an opportunity to achieve this, allowing detailed analysis of mobility patterns and enabling more effective personalized POI recommendations \citep{memon2015travel, xue2024predicting}. In this context, understanding these movement dynamics can help establish comprehensive traveler mobility paradigms for different movement contexts, which can be leveraged to recommend potential destinations that align with users' preferences and enhance their experience. By better predicting user preferences for POIs, personalized recommendation systems can significantly influence user satisfaction and shape the future of the recommendation market.

\begin{figure*}[h!]
    \centering
    \includegraphics[width=0.9\textwidth]{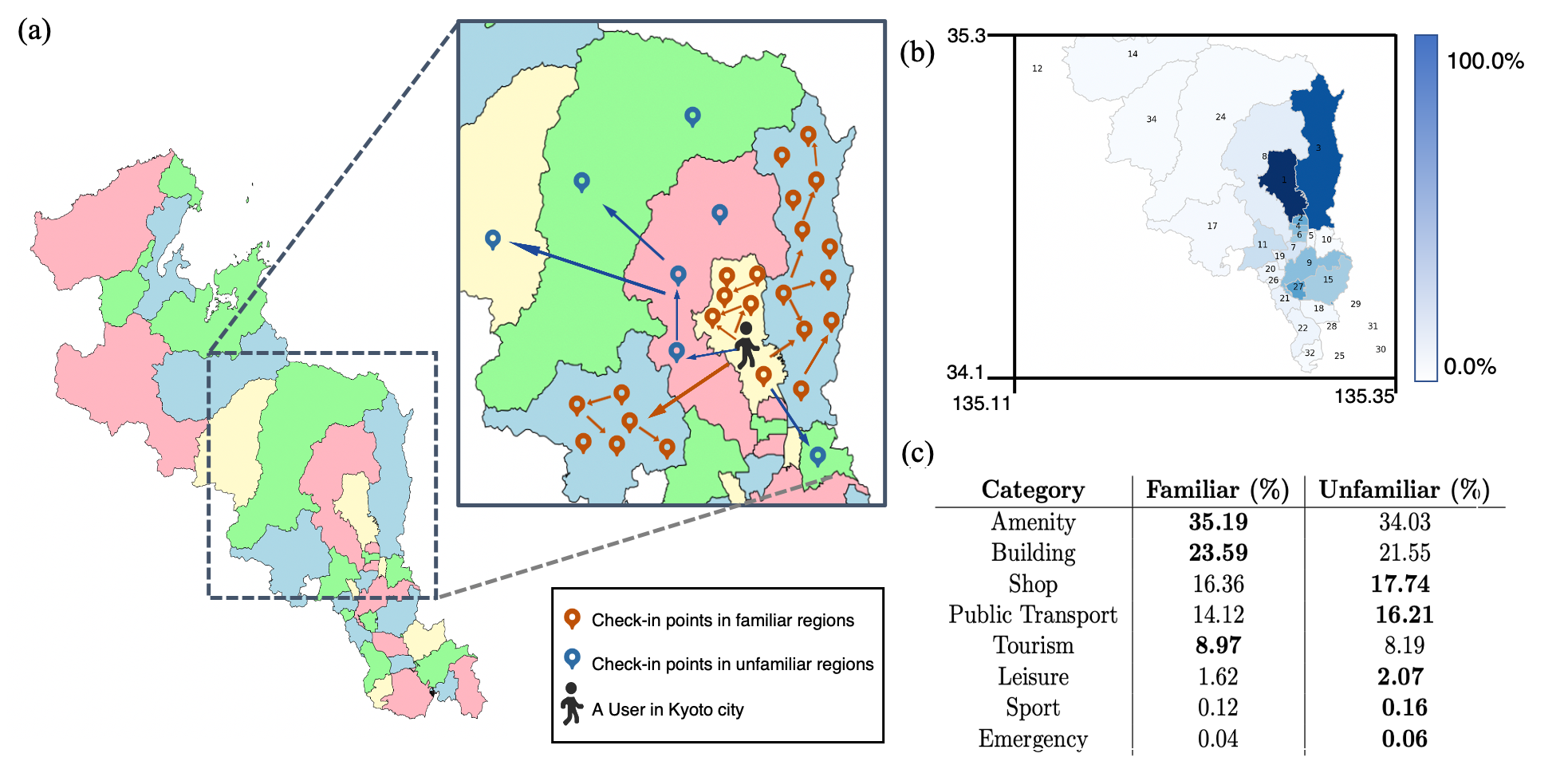} 
    \caption{Illustration of the definitions of familiar and unfamiliar movements, along with real-world examples and statistical results from the LY Corporation dataset. (a) Illustration of familiar movement and unfamiliar movement; (b) The footprint frequency demonstration of one real user in the LY Corporation dataset; (c) The visited POI types percentage difference in users' familiar and unfamiliar regions.} 
    \label{fig:figure1} 
\end{figure*}

Current methods approach the urban mobility prediction (next POI recommendation) problem as a sequential prediction task \citep{zhao2020go, wang2019urban}. These methods focus on identifying patterns in users' check-in behaviors to infer their next destinations by summarizing personalized preferences. Additionally, spatio-temporal context is often utilized to extract features of both users and POIs, aiding in the prediction process. Early studies used recurrent neural networks (RNNs) and attention mechanisms to capture user POI sequence patterns \citep{liu2022bidirectional}. Later, graph-based models were developed to construct global POI transition graphs and user similarity graphs, enhancing the representation of users and POIs \citep{yu2024survey}. More recently, Transformer models, utilizing their self-attention mechanism, have been used to deal with longer and larger data sequences to predict the next POI, achieving higher accuracy and efficiency. However, there is a limitation to these approaches. These approaches lack an in-depth analysis of providing unique POI recommendations for users in different movement contexts. Existing methods typically train a single model without considering the differences between movements in familiar and unfamiliar regions, leading to reduced recommendation accuracy in users' unfamiliar movements. Studies have shown significant differences between users who frequently travel within familiar regions and those who visit unfamiliar areas, such as travelers from other regions or countries. These differences are pronounced in various aspects, including POI type preferences and trajectory selection patterns \citep{cheung2019contingent, qiu2021valuing}. Our real-world dataset also supports this observation. As illustrated in Figure \ref{fig:figure1}(c), the category choice distribution differs between familiar and unfamiliar regions, with users exhibiting different POI preferences in each context. However, existing methods often fail to distinguish between the movements of users in their familiar and unfamiliar regions. More importantly, training a single Transformer model for all movement contexts forces the model to learn a unified mobility representation, which can lead to negative transfer when familiar and unfamiliar behaviors follow inherently different mechanisms. This motivates the need for a framework capable of capturing shared mobility patterns while simultaneously learning context-specific behaviors through adaptive specialization.

To address this problem, we propose MoE-TransMov, a Mixture-of-Experts (MoE) Transformer-based neural network for the next POI prediction that accounts for both familiar and unfamiliar movement patterns. Specifically, we first collect check-in data from a representative city in Japan and categorize user movements into familiar and unfamiliar regions to form distinct test datasets. We then introduce a Transformer-based neural network trained on the entire dataset to learn user mobility patterns and evaluate the model on the familiar and unfamiliar datasets. The core idea of MoE-TransMov is to leverage the MoE mechanism to disentangle heterogeneous mobility behaviors: the shared Transformer encoder captures global user movement structures, while multiple experts specialize in different mobility contexts. A gating network dynamically selects the most relevant experts based on a user's movement type, allowing the model to adaptively balance region-specific characteristics. This design enables the model to provide more accurate recommendations compared to a single shared model. To further assess the generalizability of our model, we additionally extract a publicly available check-in dataset from Foursquare in New York City (NYC), which is smaller in users and POI numbers compared to our primary dataset, on which we test MoE-TransMov on this external dataset to demonstrate the robustness and transferability across cities. Compared to baseline models, our approach achieves higher accuracy and efficiency, outperforming state-of-the-art methods on both familiar and unfamiliar movement datasets. Our methods for human mobility prediction have three main contributions:

\begin{itemize}

\item The study focuses on exploring the variations in human mobility patterns between familiar and unfamiliar regions of users. By capturing intrinsic differences in mobility behaviors between these regions, we improve the performance of the model in the next POI prediction;

\item We propose a novel approach designed to capture and leverage the unique characteristics of mobility in familiar and unfamiliar regions using a MoE and Transformer architecture, where the MoE structure enables adaptive expert selection to model heterogeneous movement behaviors and avoid negative transfer across regions, resulting in more precise and efficient individual movement predictions;

\item The effectiveness and generalizability of the proposed Transformer-based approach is validated using the open-sourced small dataset of NYC and a real-world check-in data from a Japanese city, highlighting its practical applicability in POI recommendation systems.
\end{itemize}

\section{RELATED WORK}

\subsection{Next POI Prediction}
The ability to predict human movements is critical to the design of travel recommendations \citep{isinkaye2015recommendation}. As shown in the Table \ref{Table1}, to develop personalized POI recommendations based on user check-in data, earlier research works on POI recommendation mainly focused on feature engineering and conventional machine learning-based methods. In this context, Markov chain based stochastic models have been explored for the next-POI prediction application \citep{gambs2012next, xia2009modelling}. With the development of matrix factorization (MF) \citep{koren2009matrix} in other recommendation systems. Then it was applied for the next POI prediction \citep{rahmani2020joint}. Later, the Bayesian network \citep{park2006context, zhang2007efficient} was introduced to achieve a better result. Traditional approaches, such as Support Vector Machine (SVM) \citep{cortes1995support} and Gaussian modeling \citep{bishop2006pattern}, have been extensively utilized in various studies for the next POI prediction. Recently, deep neural networks, including Convolutional Neural Networks (CNN) \citep{lecun1995convolutional} and Recurrent Neural Networks (RNN) \citep{hochreiter1997long}, have demonstrated significant advantages, particularly in automating feature extraction and addressing the complexities associated with manual feature engineering \citep{yao2023predicting}. Moreover, Transformer models \citep{vaswani2017attention} have the ability to capture long-range dependencies and complex patterns in user behavior sequences through self-attention mechanisms, allowing more accurate and efficient predictions. Specifically, building on the foundation laid by recurrent models, attention-based frameworks have significantly improved next POI prediction by highlighting critical spatio-temporal dependencies within user mobility data. DeepMove, for example, employs attention mechanisms to dynamically identify and prioritize important patterns in user check-in sequences, allowing the model to capture both short-term preferences and long-term behavioral trends \citep{feng2018deepmove}. STAN (Spatio-Temporal Attention Network) further extends this concept by incorporating specialized attention mechanisms that simultaneously focus on spatial and temporal dimensions, ensuring that key features are emphasized during the prediction process \citep{luo2018stan}. 

\begin{table}[t]
\small
\centering
\footnotesize
\caption{Comparison of Representative Next POI Prediction Methods}
\label{Table1}
\renewcommand{\arraystretch}{1.25}
\setlength{\tabcolsep}{5pt}

\resizebox{\textwidth}{!}{
\begin{tabular}{p{2.4cm} p{2.8cm} p{2.0cm} p{2.3cm} p{3.4cm} p{1.2cm} p{3.4cm} p{2cm}}
\toprule
\textbf{Category} & \textbf{Paper} & \textbf{Model Type} & \textbf{Dataset} & \textbf{Key Mechanism} & \textbf{Spatial Modeling} & \textbf{Temporal Modeling} & \textbf{User Preference} \\
\midrule

\multirow{3}{*}{\textbf{Traditional}}
& Markov \citep{gambs2012next,xia2009modelling} & Probabilistic Chain & Foursquare, Gowalla & 1st-order transition probability & No & No & No \\

& MF \citep{koren2009matrix,rahmani2020joint} & Latent Factorization & Foursquare, Gowalla & User--POI latent factors & Weak & No & Limited user representation \\

& Bayesian \citep{park2006context,zhang2007efficient} & Probabilistic Graph & Early POI logs & Conditional dependency modeling & No & Weak sequential & No \\
\midrule

\multirow{2}{*}{\textbf{RNN / CNN}}
& DeepMove \citep{feng2018deepmove} & RNN + Attention & Foursquare NYC, TKY & Short-term attention + periodicity & No & Short-sequence modeling & No \\

& STAN \citep{luo2018stan} & ST-Attention Network & Foursquare NYC, Gowalla & Spatial + temporal attention & Yes & Yes & Limited user representation \\
\midrule

\multirow{2}{*}{\textbf{Transformer}}
& Transformer \citep{vaswani2017attention} & Self-attention Encoder & Foursquare, Gowalla & Multi-head attention & Yes & No & Limited user representation \\

& Informer \citep{zhou2021informer} & Efficient Transformer & ETTh/ETTm, Electricity, Weather & ProbSparse attention & No & Long-sequence modeling & No \\
\midrule

\multirow{3}{*}{\textbf{Graph-based}}
& GETNext \citep{yao2020getnext} & Graph Sequential Model & Foursquare NYC, TKY & Trajectory flow + POI graph & Yes & Short-sequence modeling & Limited user represen-
tation \\

& TrajGraph \citep{zhao2024trajgraph} & Graph Transformer & Foursquare, Gowalla, Geolife & Dual-view ST graph & Yes & Yes & No \\

& LGSA \citep{zeng2023lgsa} & Graph Attention Network & Foursquare, Gowalla & Local--global attention & Yes & Yes & No \\
\midrule

\multirow{2}{*}{\textbf{MoE-based}}
& ST-MoGE \citep{wu2024st_moge} & Graph MoE & Gowalla, Foursquare NYC & MoE + GNN for ST patterns & Yes & Yes & Limited user representation \\

& PTBformer-MMoE \citep{xi2025multi} & MoE + Transformer & go-card (Australia) & Multi-mode Transformer + MMoE & Yes & Long-sequence modeling & Passenger-type aware \\
\midrule

\textbf{Our Proposed Model}
& \textbf{MoE-TransMov}
& \textbf{MoE + Transformer}
& \textbf{Foursquare NYC, Japan LY dataset}
& \textbf{Experts routing for different movement types}
& \textbf{No}
& \textbf{Long-sequence \& short-sequence modeling}
& \textbf{User familiarity modeling} \\

\bottomrule
\end{tabular}}
\end{table}

Recent advancements in POI recommendation have increasingly leveraged Transformer-based and graph-based models to capture complex spatial-temporal dependencies in user mobility. Transformer-based models, such as Informer, have demonstrated significant efficacy in modeling long-range dependencies and dynamic mobility contexts. Informer introduces a ProbSparse self-attention mechanism, which reduces the time complexity from quadratic to logarithmic scale, thereby enabling efficient processing of long POI sequences \citep{zhou2021informer}. This efficiency is particularly beneficial in handling extensive user check-in data, leading to more accurate next-POI predictions. On the other hand, graph-based models such as GETNext and TrajGraph represent user movements as structured networks, where nodes correspond to POIs, and edges denote transitions between them. GETNext uses a user-agnostic global trajectory flow map, integrating global transition patterns, user preferences, and spatio-temporal contexts to improve prediction accuracy \citep{yao2020getnext}. Similarly, TrajGraph utilizes a dual-view graph transformer model to capture both spatial and temporal dependencies in user trajectories, effectively modeling the complex relationships between POIs \citep{zhao2024trajgraph}. These models exemplify the growing emphasis on understanding the interplay between time and space in user trajectories.

Moreover, user interest is also an important factor in the next POI prediction study. To incorporate user interests more comprehensively, \cite{zeng2023lgsa} proposed LGSA, a method combining local and global preferences while considering temporal and spatial awareness for improved POI selection. Graph-based solutions are another emerging line of research \citep{ huang2024jointly}. \cite{yin2023next} designed a dynamic graph model with explicit dependency modeling to capture temporal and semantic relations between POIs, showing strong adaptability to evolving user behavior. Furthermore, to address the challenges of data sparsity and generalization in cross-city scenarios, \cite{xu2024crosspred} developed CrossPred, a framework that leverages POI feature matching across cities to predict long-distance traveler behavior. These studies collectively demonstrate the importance of integrating temporal dynamics, spatial context, user preferences, and domain-specific knowledge in advancing next POI prediction models.

\subsection{MoE Framework}

The MoE framework has emerged as a powerful approach for handling diverse and complex patterns in sequential data. By dynamically selecting specialized sub-models (experts). For each input, MoE enables the model to learn distinct patterns within subpopulations of data, thus achieving greater representation capacity and flexibility \citep{jacobs1991adaptive,shazeer2017outrageously}. This activation mechanism enables the model to selectively focus on different patterns inherent in sequential POI data, broader trends, and movements in their familiar or unfamiliar regions \citep{zhou2022mixture,wu2024st_moge}. The gate typically employs a Softmax function to route inputs to the most relevant experts, enhancing the expressiveness of the model without proportionally increasing computational costs \citep{shazeer2017outrageously}. As a result, despite having billions of parameters, only a small portion is active during each forward pass, enabling the model to handle complex sequences efficiently \citep{lepikhin2020gshard}. In the context of the next POI recommendation, MoE can be particularly effective in addressing the heterogeneity of user mobility behaviors. For example, by assigning different experts to users with familiar versus unfamiliar movement patterns, the framework can capture nuanced travel behaviors and preferences unique to each subgroup \citep{wu2024st_moge}. Furthermore, the gating mechanism in MoE ensures computational efficiency by activating only a subset of experts for each prediction, making it scalable to large datasets. Recently, some studies used MoE on the human mobility study. One of the study focus on the spatial-temporal mix of graph experts (ST-MoGE) framework demonstrates the adaptability of MoE to heterogeneous data through the use of multiple specialized submodels \citep{wu2024st_moge}. Another study \citep{xi2025multi} introduced PTBformer-MMoE, a multi-task Transformer framework that leverages multi-gate MoE to jointly model periodic travel behavior and personalized fare patterns in large-scale multimodal public transport systems, demonstrating the effectiveness of expert specialization in capturing heterogeneous traveler behaviors. However, according to some studies, integrating MoE into the next POI prediction models presents challenges such as expert redundancy and imbalance in expert utilization, which require careful design of the gating network and optimization strategies to ensure robust and interpretable predictions for the next POI prediction tasks \citep{shazeer2017outrageously,wu2024st_moge}.

\subsection{Transformer Models}

Transformer models consist of self-attention mechanism and parallelizable architecture, which excel in capturing long-range dependencies within user movement trajectories by assigning dynamic attention weights to all points in a sequence, irrespective of their distance \citep{vaswani2017attention}. This capability allows the model to effectively capture both short-term preferences and long-term mobility trends. Additionally, the inclusion of positional embeddings ensures that the temporal ordering of POIs is preserved, addressing a key challenge in mobility sequence modeling. Recent studies have extended Transformers to integrate multi-modal and heterogeneous data, such as combining several different check-in trajectory patterns. This integration enables the model to consider both movement patterns and contextual information, enhancing its prediction accuracy. For example, hybrid Transformer architectures, such as the MoE Transformer, dynamically allocate specialized sub-networks to handle diverse data sources, making them particularly effective for capturing the complex interplay between movements of users' familiar and unfamiliar regions in urban environments.

Despite the fact that transformer architectures have proven highly effective, scalability issues due to increasing computational and memory requirements as model sizes increase remain a challenge \citep{khan2022transformers}. A promising solution to this challenge is the MoE approach, which can significantly expand the capacity of the model while maintaining computational efficiency \citep{wang2020deep}. To address this problem, using MoE in this Transformer context, the model gains a unique ability to handle diverse POI sequences, learning more nuanced representations that improve prediction accuracy over standard dense Transformer models. The integration of MoE into the Transformer architectures provides an efficient and scalable method for next POI prediction, offering the dual benefits of enhanced capacity and reduced computational overhead. This approach not only improves the ability to capture complex movement behaviors, but also allows for the efficient deployment of models on large-scale datasets typical in this domain.

\section{PRELIMINARIES}

In this section, we introduce the detailed concepts within our problem definition.

\textbf{Definition 1: (POI):} A POI is a spatial site (e.g., a park, a shopping mall, etc.) associated with three attributes: a unique identifier, geographical coordinates, and the types of POI tuples, which can be denoted as $q_i:$ ($\sigma, x, y, z$), where $\sigma$ is the unique id and  $x, y$ are the longitude and latitude, $z$ is the type of POI (Amenity, Shop, Public Transportation, etc.) 

\textbf{Definition 2: (Trajectory Sequence)} Given a user identification $u$, trajectory sequence $S$ is a POI sequence, i.e., $S^u = q_1 q_2 \cdots q_n$, where $q_i$ refers to POIs for check-in points within the user's movement trajectories. 

\textbf{Definition 3: (Main Activity Regions)} The main activity region is defined as the region that is the center of the cluster of each user's daily check-in points within the first weekday of check-in records of a user.

\textbf{Definition 4: (Familiar and Unfamiliar Movements)} Familiar movements are defined as check-in movements that occur within a user's familiar region, while unfamiliar movements refer to check-in movements that take place outside this region. In this study, the three most frequently visited regions of each individual as well as their main activity regions, are classified as familiar regions. The movements within these regions are defined as familiar movements. All other movements are considered unfamiliar movements. The patterns of the two movements are demonstrated in Figures \ref{fig:figure1}(a) and \ref{fig:figure1}(b). 

\textbf{Problem 1: (Next POI Prediction)} The objective of the next POI prediction: Given the current trajectory $S^u = q_1 q_2 \cdots q_n$ and the corresponding trajectory history $S'^u = q_1 q_2 \cdots q_{n-1}$, predict the next POI $q_{n}$ in the trajectory and list of top-ranked next POIs that he/she may be most likely in the next time is used to evaluate the performance. At each prediction step, the model only has access to a user’s observed trajectory history prior to the prediction time. The target POI $q_n$ will never be included in the input sequence during either training or inference. 

\begin{table}[ht]
\scriptsize
\centering
\caption{Definition of Parameters}
\label{Table2}
\begin{tabular}{@{}p{2cm}p{6cm}@{}}
\toprule
\textbf{Symbol} & \textbf{Description} \\
\midrule
$q_i$           & A POI represented as a tuple $(\sigma, x, y, z)$ \\
$\sigma$        & Unique identifier of a POI \\
$x, y$          & Longitude and latitude of a POI \\
$z$             & Type of POI (e.g., Amenity, Shop, Public Transportation, etc.) \\
$u$             & User identifier \\
$S^u$           & A user’s full trajectory sequence: $q_1 q_2 \cdots q_n$ \\
$S'^u$          & A user’s historical trajectory sequence: $q_1 q_2 \cdots q_{n-1}$ \\
\bottomrule
\end{tabular}
\label{tab:notations}
\end{table}

\section{THE MoE-TransMov MODEL}

\begin{figure*}[h!]
    \centering
    \includegraphics[width=0.96\textwidth]{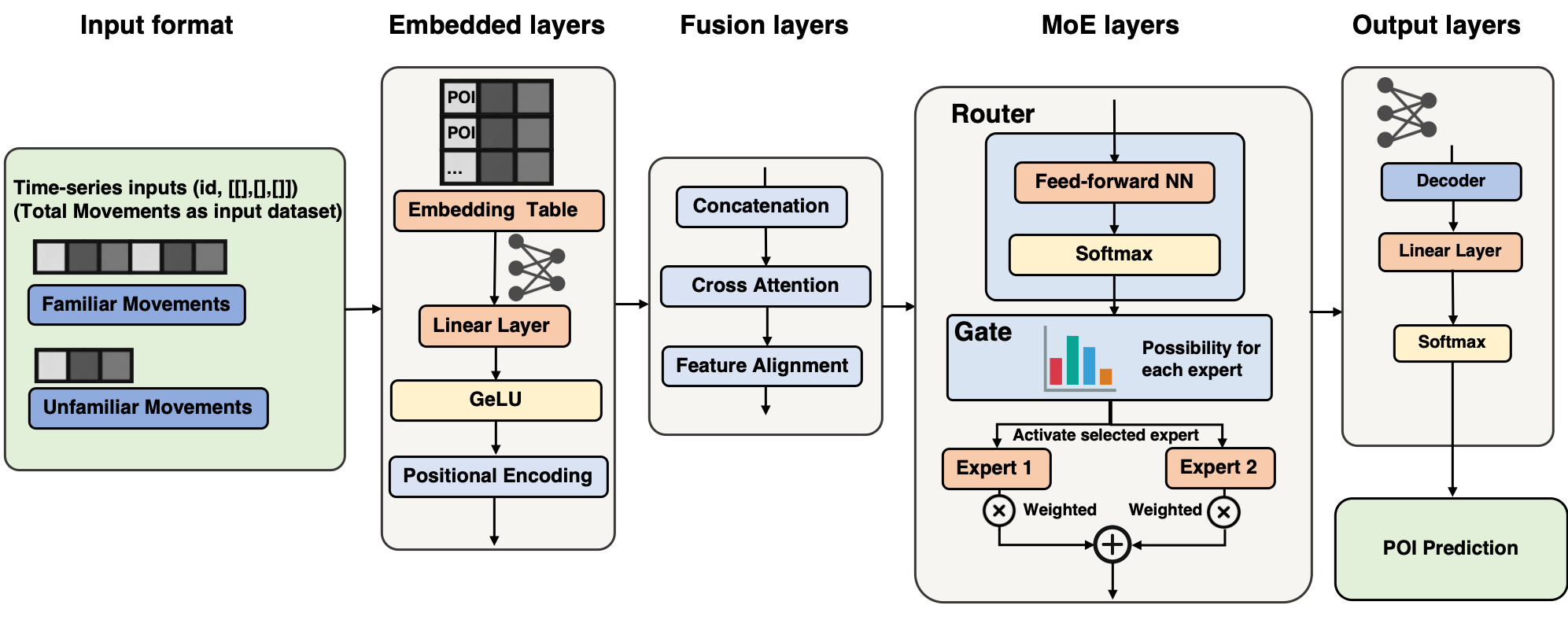} 
    \caption{Framework Overview of Our Proposed MoE-TransMov Model: Embedding, Fusion, and MoE Layers} 
    \label{fig:figure2} 
\end{figure*}

In Figure \ref{fig:figure2}, we demonstrate the overall design of our proposed MoE-TransMov model. The model takes sequences that may belong to two movement contexts: (1) familiar movements, representing the mobility patterns of the time series of users within familiar regions; (2) unfamiliar movements, capturing the mobility patterns of users in unfamiliar regions. These inputs allow the model to take advantage of both spatial and contextual factors for an accurate POI prediction.

\subsection{Embedding and Fusion Layers}
To process the input data, we designed an embedding table for each POI, and each POI has $d$ trainable factors in that table to reduce the influence of the POI index, since the POI is represented in continuous number format. Different input is passed through a Linear Layer to project it into a higher-dimensional space, enhancing the feature representation. A GeLU activation function is applied to introduce non-linearity, ensuring stable gradient propagation.

Before the MoE layers, we employ an attention mechanism. This structure reduces redundancy while maintaining the ability to capture complex dependencies in sequential data. The attention mechanism allows the model to focus on relevant POI transitions by learning weighted relationships across different input sequences, enhancing the model’s ability to generalize between familiar and unfamiliar regions. In addition, layer normalization and residual connections are applied to stabilize training and improve information flow within the network.

\subsection{MoE Layers}

In these layers, the gating network does not take the familiar/unfamiliar label as input during either training or inference. Expert routing is learned solely from the historical trajectory representation via the gating network, without using any future movement information. As shown in Algorithm \ref{alg:1} and Figure \ref{fig:figure2}, we define two experts in the MoE layers for predicting familiar and unfamiliar movements. This MoE model for the next POI prediction first computes gating weights by applying a softmax function to the gating logits, determining the contribution of each expert. Based on the gating probabilities, the model adaptively weighs multiple expert networks, each corresponding to a different inductive bias for modeling mobility sequences: LSTM-based experts with capacity of dealing short POI sequences and Transformer-based experts with capacity of dealing long sequences. Each expert processes the input separately, and their outputs are combined using a weighted sum derived from the gating mechanism.

\begin{algorithm}[H]
\caption{MoE Layers for Next POI Prediction}
\label{alg:1}
\begin{algorithmic}[1]
\small
\Require Input POI sequence $\mathbf{x} = [\text{POI}_1, \text{POI}_2, \dots, \text{POI}_T]$, number of experts $N$
\Ensure Predicted next POI

\State \textbf{Step 1: Compute Gating Weights}
\Statex \hspace{1em} Compute gating logits: $\mathbf{g}_{\text{logits}} = \mathbf{W}_g \cdot \mathbf{x} + \mathbf{b}_g$
\Statex \hspace{1em}Apply softmax to obtain gating probabilities: 
\[
\mathbf{g} = \text{Softmax}(\mathbf{g}_{\text{logits}})
\]
\State \textbf{Step 2: Select and Compute Expert Outputs Based on Gating network results}

\State \textbf{Expert 1 (For unfamiliar movements)}
    \Statex \hspace{1em} Use experts with short and sparse sequences modeling capacity:

        \State Compute output for unfamiliar-data expert $i$:
        \[
        \mathbf{y}_{\text{expert}, i} = \text{LSTM}_{i}(\mathbf{x})
        \]

\State \textbf{Expert 2 (For familiar movements)}
    \Statex \hspace{1em} Use experts with strong long-range sequence modeling capacity
        \State Compute output for Familiar-data expert $i$:
        \[
        \mathbf{y}_{\text{expert}, i} = \text{Transformer}_{i}(\mathbf{x})
        \]

\State \textbf{Step 3: Weight Experts and Combine Outputs}

\State Compute weighted sum of experts:
\[
\mathbf{y}_{\text{output}} = \sum_{i=1}^N g_i \cdot \mathbf{y}_{\text{expert}, i}
\]

\State \textbf{Step 4: Predict the Next POI}
\Statex \hspace{1em} Apply a decoder to map the combined output to the next POI:
\[
\hat{\text{POI}} = \text{Decoder}(\mathbf{y}_{\text{output}})
\]
\end{algorithmic}
\end{algorithm}

In the MoE structure, we define the gating logits as below:
\begin{equation}
    g_{\text{logits}} = W_g \cdot x + b_g
\end{equation}
where:
\begin{itemize}
    \item \( x \) is a representation of the current sequence and its final embedding,
    \item \( W_g \) and \( b_g \) are trainable parameters.
\end{itemize}

Then, we apply Softmax to obtain gating probabilities:
\begin{equation}
    g = \text{Softmax}(g_{\text{logits}})
\end{equation}
Where vector \( g \) will have as many components as there are experts \( N \). In this particular setup, \( N = 2 \). One for LSTM, and one for Transformer. Thus,
\begin{equation}
    g = [g_1, g_2],  g_1 + g_2 = 1.
\end{equation}

We chose LSTM and Transformer because of their features that LSTMs are good at handling short sequences. They do not rely heavily on longer‐range context and have comparatively fewer parameters than Transformers, making them more suitable when data is sparse or the sequence length is limited, while Transformers excel at modeling long‐range dependencies due to self‐attention. Transformer can leverage their attention mechanism to capture complex, global dependencies more effectively than a purely recurrent model. 

The gating probabilities determine how much each expert contributes to the combined output. After each expert produces its output, these outputs are combined by a weighted sum driven by \( g \):
\begin{equation}
    y_{\text{output}} = \sum_{i=1}^{N} g_i \cdot y_{\text{expert}, i}.
\end{equation}

Finally, a decoder maps the aggregated output to predict the next POI that users might visit.

\subsection{Loss Function and Output Layers}
Regarding the loss function and output layers chosen for this model, all MLP heads in the decoder are utilized during training, with the output serving as regularization components. This means that we compute a weighted aggregate of the losses from all MLP heads. For POI and POI category prediction, cross-entropy is employed as the loss function. 

Let \(M\) denote the number of training samples and \(C\) represent the total number of unique POIs. For each sample \(i\), the model outputs a set of normalized scores \(\{z_{i,1}, z_{i,2}, \dots, z_{i,C}\}\), where each \(z_{i,c}\) represents the likelihood of the next POI being POI \(c\). To obtain a valid probability distribution over all possible POIs, we apply the softmax function:

\begin{equation}
\hat{p}_{i,c} = \frac{\exp(z_{i,c})}{\sum_{j=1}^{C} \exp(z_{i,j})},
\end{equation}

where \(\hat{p}_{i,c}\) represents the predicted probability that the next POI for the \(i\)-th sample belongs to POI \(c\).

The cross-entropy is used as the loss function in the following format. Given the true next POI label \(y_i \in \{1,2,\dots,C\}\), the cross-entropy loss for one sample is defined as:

\begin{equation}
\ell_i = -\log \bigl(\hat{p}_{i,y_i}\bigr).
\end{equation}

By averaging this loss over all \(M\) training samples, we obtain the total loss function:

\begin{equation}
\mathcal{L}_{\mathrm{CE}} 
= \frac{1}{M} \sum_{i=1}^{M} \ell_i 
= - \frac{1}{M} \sum_{i=1}^{M} \log \bigl(\hat{p}_{i,y_i}\bigr).
\end{equation}

The processed features are passed to the Output Layers, where a decoder extracts high-level task-relevant information. A Linear Layer projects the decoded features into the target space, and the Softmax function generates a probability distribution over the potential POI predictions. This hierarchical design ensures that the model effectively maps the fused inputs to accurate predictions. The final output of the model is the predicted POI. By integrating information from data from familiar and unfamiliar mobility patterns, the model will be able to achieve robust prediction performance, capturing the complexity of user movement and behavior patterns.

\section{DATA DESCRIPTION}

\subsection{Check-in and Main Activity Region Data}

In this study, we use two datasets to comprehensively evaluate the performance and generalizability of our proposed model. We first validate the model using the publicly available Foursquare NYC dataset, a widely adopted benchmark in next POI prediction research. This open-source dataset allows fair comparison with existing methods and provides a standardized evaluation environment. After establishing the baseline performance, we further train and test our model on a large-scale real-world dataset collected in Japan by LY Corporation. This proprietary dataset contains significantly more POIs, denser check-in behaviors, and richer mobility patterns, enabling us to assess the robustness and practical applicability of our model in realistic urban scenarios. The datasets used in this paper contains the anonymized user ID, the location of the POIs that users visited during a certain period, the user's check-in timestamp, and the approximate main activity region of the users. The data were collected by our collaborator LY Corporation. The data we chose in this study include POI check-in data from Kyoto City, Japan. The period we chose is $01/01/2023 - 09/01/2023$, which contains user check-in data of 8 months and more than one million time sequences based on the user's trajectory during daily life. 

\textbf{Check-in Data:} The collected dataset includes data formats such as anonymized user ID, POI visited timestamps, geographic longitudes and latitudes of their POIs visited. The frequency of data collection was around two hours for each interval. In this study, we used the data to predict the next POI that users will visit and check in. Before the experiment, we filter out sessions with fewer than 10 records and users with fewer than 10 sessions. 
We only use POI information from daily check-in of users and do not use private information such as home or office for our analysis.

\textbf{Main Activity Region Data:} The main activity region represents the geographic center of the user's daily movements. It is determined based on their check-in POI locations. Specifically, we identify this region by computing the cluster center of all check-in points recorded within the user's first seven days of activity. The region containing this cluster center is designated as the user's main activity region.

\subsection{Data Pre-processing}

The latitude and longitude of the POIs were first down-sampled and then used to estimate their main activity region, the POIs they have visited. Data from the main activity region are used to estimate the activity zones of the users. 

\textbf{Main Activity Region Estimation}: To infer the most likely location of a user's primary activity region, we utilize check-in data collected over the whole day (24 hours) during the first weekday. The mean shift algorithm \citep{carreira2015review} is recognized for its robustness and efficiency in clustering, excels at pinpointing dense areas of data points. When applied to check-in data, this algorithm facilitates the identification of the primary activity region. Then, we identify the regions a user visits most by computing the share of visits each region accounts for among the user’s total movements. Together, these identified regions are defined and utilized as a feature to classify users as familiar movements or unfamiliar movements in the area.

\textbf{Familiar and Unfamiliar Movements Classification}: Figure \ref{fig:figure1} (a) illustrates the distinction between familiar and unfamiliar movements. We classify each user's activity region into two categories: familiar activity region and unfamiliar activity region. A familiar activity region is defined as the region that contains the top three regions where the user has made the highest number of visits, accounting for their most frequent movements. In contrast, an unfamiliar activity region encompasses all regions outside of these regions.

\section{EXPERIMENT AND RESULTS}
\subsection{Setting, Evaluation Metrics, Baselines}
In our experiments, the baseline and proposed models are trained using an A100 GPU and we adopt an early stop strategy that terminates training if the loss increases for three consecutive epochs. There are two experts in total: one specialized for familiar movements and the other for unfamiliar ones. To benchmark our model, we implement several existing prediction models: 
\begin{itemize}
    \item Majority Vote, a basic statistical model that predicts the next POI by selecting the most frequently visited location from historical data \citep{penrose1946elementary};
    \item MLP, a feedforward neural network that captures non-linear relationships in POI sequences. It processes input features through multiple layers of fully connected neurons with activation functions, enabling it to learn complex patterns \citep{pinkus1999approximation};
    \item LSTM, a type of recurrent neural network designed to handle long-term dependencies in sequence data \citep{hochreiter1997long};
    \item DeepMove, a deep learning model that employs attention mechanisms to capture spatiotemporal patterns in user check-in sequences \citep{feng2018deepmove}; 
    \item STAN, a POI prediction model that enhances prediction by focusing on important spatial and temporal aspects \citep{luo2018stan}; 
    \item GETNext, a state-of-the-art POI prediction model that integrates both geographical and temporal contexts, which utilizes graph-based structures and sequence modeling to enhance user trajectory understanding, enabling it to make more contextually aware predictions \citep{yao2020getnext};
    \item Transformer, a self-attention-based deep learning model that eliminates recurrence and processes entire sequences in parallel. Transformers efficiently capture global dependencies in POI sequences by learning relationships between distant locations \citep{vaswani2017attention};
    \item Informer, an efficient long-sequence time-series forecasting model that leverages the ProbSparse self-attention mechanism to reduce computational complexity, enabling fast and accurate predictions with high scalability \citep{zhou2021informer}.

\end{itemize}

To evaluate the effectiveness and generalizability of our proposed model, we first conduct experiments on the widely used open-source Foursquare NYC dataset. This dataset provides a standardized benchmark and enables a fair comparison with existing state-of-the-art approaches. After validating the model’s baseline performance, we further train and test it on our proprietary LY dataset collected in Japan. Compared to the Foursquare dataset, the LY dataset contains approximately five times more unique POIs and nearly fifty times more training samples, offering a substantially richer and more diverse mobility environment. More importantly, the LY dataset reflects real-world user behaviors and region-specific movement patterns in Japan, allowing us to assess the model’s robustness and practical applicability in realistic urban scenarios.

For all experiments, user trajectories are treated as temporally ordered sequences and the model is trained to predict next POI using only the historical observations. We split the total movement dataset into 80$\%$ training and 20$\%$ validation at the trajectory level and used a set of hyperparameter to get optimal performance. For the structure of our proposed model, an expert with strong long-range sequence modeling capacity (Transformer-based) is designed for familiar movements with four encoder layers, multi-head attention has 8 heads, and the feed-forward layers with hidden dimensions of 128. Expert biased toward short and sparse sequences (LSTM-based): We designed a two-layer LSTM with hidden size of 64. As shown in both the code and the architecture diagram (Figure \ref{fig:figure2}), there are two experts and the router network learns a two-class softmax to determine gating probabilities [g$_1$, g$_2$], dynamically weighting the Transformer and LSTM experts. The model processes sequences of length 50 for training and supports a maximum sequence length of 500 for positional embeddings. The training setup uses a batch size of 64, an Adam optimizer with a learning rate of 0.0005, where we tried the learning rate of 0.001, 0.0005, and 0.0001 and we finally chose 0.0005 as our learning rate.

For evaluation, we assessed the models trained on the total movement dataset by measuring POI prediction performance using Top-k precision (k = 1, 5, 10) and mean reciprocal rank (MRR) on the familiar, unfamiliar, and total datasets. Specifically, we rank the predicted probabilities of all POIs from highest to lowest. Then, we check if the actual visited POI appears among the Top-k predicted POIs. Then we also compare the MRR of the prediction with each group, which MRR is applied to evaluate the performance of POI places prediction, which measures the quality of ranked predictions by evaluating how early the correct POI appears in the predicted list.
\begin{equation}
    MRR = \frac{1}{|Q|} \sum_{i=1}^{|Q|} \frac{1}{rank_i}
\end{equation}

where:
\begin{itemize}
    \item $|Q|$ denotes the total number of test samples.
    \item $rank_i$ represents the rank position of the first POI correctly predicted for the $i$-th sample.
\end{itemize}

\subsection{Performance Analysis}

\captionsetup{font=normalsize}


\begin{figure*}[h!]
    \centering
    \includegraphics[
        width=\textwidth,
        trim=10 10 10 10,
        clip
    ]{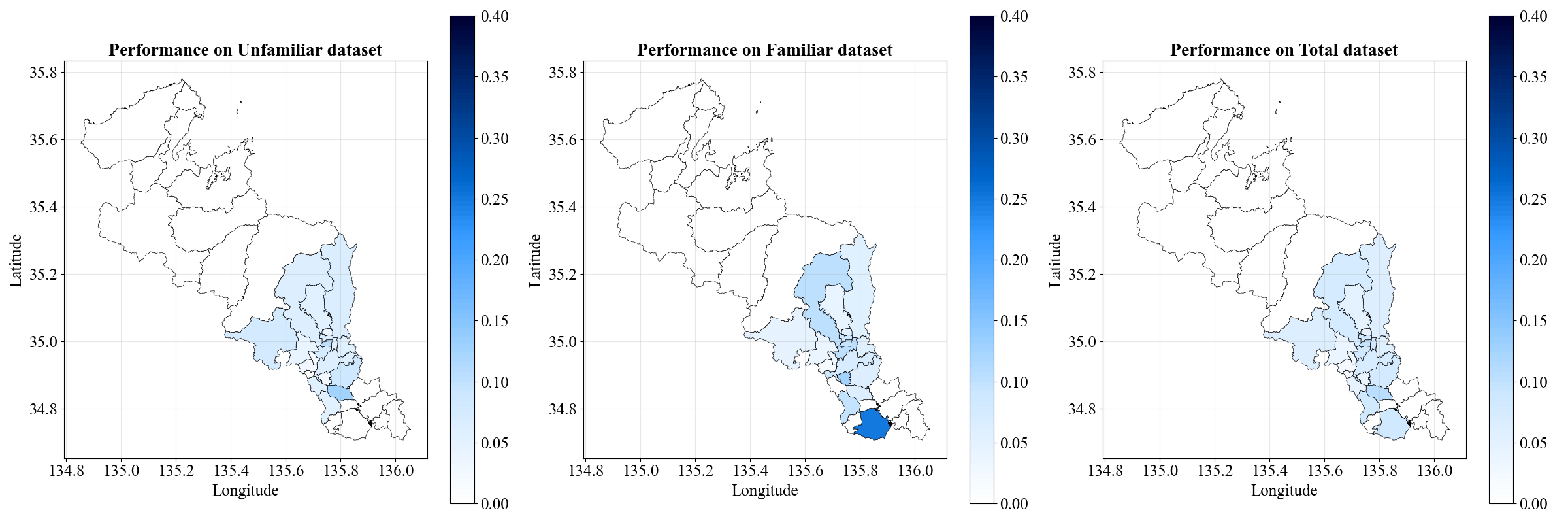}
    \caption{Spatial distribution of the Top-1 next POI prediction performance of the MoE-TransMov model across different regions under three evaluation settings (Unfamiliar, Familiar, and All) on the Kyoto Mobility Dataset. Darker shades indicate higher Top-1 prediction accuracy, while lighter shades represent lower accuracy.}
    \label{figure3}
\end{figure*}

As shown in Figure \ref{figure3}, the proposed MoE-TransMov model was applied to evaluate next-POI prediction performance across different areas of Kyoto. The results are visualized as heatmaps, where darker regions represent higher accuracy in predicting the next point of interest. The overall spatial distribution reveals that prediction performance is generally higher in the central metropolitan zones, indicating that dense urban movement patterns are better captured by the model due to the abundance and regularity of mobility data in these areas.

In the Familiar dataset, the highest accuracy is observed in the Kizugawa region, a predominantly residential area. This suggests that the movements of familiar users are more predictable, as a large proportion of residents regularly commute between Kizugawa and the metropolitan center. The homogeneous and repeated travel behaviors in this region enable the model to learn stable spatial-temporal features effectively.

For the Unfamiliar dataset, the prediction accuracy is more spatially dispersed, with no dominant high-performance cluster. This pattern reflects the heterogeneous and irregular trajectories of unfamiliar users, whose mobility behaviors are less consistent and more exploratory. Interestingly, peripheral suburban areas show slightly higher accuracy than central zones, possibly because occasional visitors tend to follow major transportation corridors or landmark-based routes that are easier for the model to capture.
When combining all datasets, the resulting heatmap shows a balanced spatial distribution, dominated by the same high-accuracy corridor from southern to central Kyoto. This indicates that the model successfully generalizes across both local and non-local movement patterns, demonstrating robust spatial learning capability across diverse urban contexts.

\subsection{Comparisons with Baseline Models}

To evaluate the performance of the baseline models, we trained all models on the Total Movement dataset and assessed their performance on three subsets: Familiar Movement, Unfamiliar Movement, and Total Movement. These subsets represent different user behavior patterns. The familiar subset contains movements users have frequently visited, while the unfamiliar subset includes new or less-visited locations. The total dataset provides a comprehensive view of all movement records.

\subsubsection{Foursquare NYC Dataset}

\begin{table}[hbtp]
\centering
\caption{Performance comparison in Top-k accuracy and MRR on Foursquare datasets}
\label{Table3}
\resizebox{\linewidth}{!}{
\begin{tabular}{lcccc|cccc|cccc}
\hline
                 & \multicolumn{4}{c|}{Familiar Movement}          & \multicolumn{4}{c|}{Unfamiliar Movement}       & \multicolumn{4}{c}{Total Movement} \\ \cline{2-13} 
                 & Top-1 & Top-5 & Top-10 & MRR   & Top-1 & Top-5 & Top-10 & MRR   & Top-1 & Top-5 & Top-10 & MRR   \\\hline
Majority Vote         &    \textbf{0.2248}    &   \textbf{0.4755}     &  \textbf{0.5698}     &   \textbf{0.3328}     &   0.0056     &  0.0893      &   0.2124     &   0.0461     &    0.1161    &   0.2839     &   0.3925     &    0.1906    \\
MLP          &    0.1278    &   0.2278     &  0.2667     &   0.1687     &   0.0423     &  0.1376      &   0.1640     &   0.0810     &    0.1023    &   0.2093     &   0.2535     &    0.1490    \\
LSTM              &    0.1639   &   0.3056     &  0.3722     &   0.2289     &   0.0952     &  0.2169      &   0.2487     &   0.1510     &    0.1535    &   0.3233     &   0.3837     &    0.2290    \\
DeepMove              &    0.1803    &   0.3127     &  0.3690     &   0.2397     &   0.0955     &  0.1910      &   0.2472     &   0.1408     &    0.1756    &   0.3044     &   0.3443     &    0.2316    \\
STAN              &    0.1472    &   0.2944     &  0.3694     &   0.2117     &   0.1005     &  0.1958      &   0.2381     &   0.1417     &    0.1581    &   0.2884     &   0.3279     &    0.2300   \\
Transformer              &    0.1389    &   0.2722     &  0.3194     &   0.1927     &   0.0476     &  0.1376      &   0.1852     &   0.0865     &    0.1209    &   0.2535     &   0.3186     &    0.1366    \\
Informer                &    0.1639    &   0.3167     &  0.3528     &   0.2228     &   0.0794     &  0.1534      &   0.2011     &   0.1106     &    0.1302    &   0.3316     &   0.3628     &    0.2078    \\
MoE-TransMov         &    0.2111    &   0.3556     &  0.4056     &   0.2750     &   \textbf{0.1164}     &  \textbf{0.2275}      &   \textbf{0.2593}     &   \textbf{0.1595 }    &    \textbf{0.1930}    &   \textbf{0.3558}     &   \textbf{0.3953}     &    \textbf{0.2639}\\
\hline
\end{tabular}%
}
\end{table}

Table \ref{Table3} summarizes the experimental results on the Foursquare dataset. Overall, MoE-TransMov shows clear superiority on the Unfamiliar Movement and Total Movement subsets, consistently outperforming all baseline methods across Top-k and MRR. This indicates that the proposed mixture-of-experts architecture effectively captures diverse mobility patterns and provides strong generalization when encountering sparse POIs. 

Interestingly, on the Familiar Movement subset, the Majority Vote baseline achieves the highest Top-1, Top-5, and Top-10 accuracy, with values of 0.2248, 0.4755, and 0.5698, respectively, outperforming learning-based models including MoE-TransMov (Top-1: 0.2111, Top-5: 0.3556, Top-10: 0.4056). This result can be largely attributed to the inherent characteristics of the Foursquare dataset, where user mobility is highly skewed toward a small number of frequently visited POIs, leading to strong repeat-visit and popularity-dominated patterns. Under such dataset conditions, users tend to revisit habitual locations, making simple frequency-based heuristics highly competitive for predicting familiar movements. However, this dataset-specific advantage does not generalize to unfamiliar or total movement contexts. On the Unfamiliar Movement subset, Majority Vote performs poorly (Top-1: 0.0056, Top-5: 0.0893, Top-10: 0.2124), whereas MoE-TransMov achieves substantially higher accuracy (Top-1: 0.1164, Top-5: 0.2275, Top-10: 0.2593) and a markedly improved MRR (0.1595 vs. 0.0461). More importantly, when evaluated on our large-scale real-world dataset collected by LY Corporation, which exhibits richer POI diversity, longer trajectories, and less extreme popularity concentration, MoE-TransMov consistently outperforms all frequency-based and learning-based baselines across both familiar and unfamiliar movement scenarios. These results indicate that while popularity-driven baselines may benefit from the specific structural properties of smaller and highly skewed datasets such as Foursquare, the proposed MoE-TransMov model demonstrates stronger robustness and generalization by capturing heterogeneous mobility patterns beyond frequency-based prediction, particularly in realistic and data-diverse settings.

\subsubsection{LY Corporation Dataset}

\begin{table*}[hbtp]
\centering
\caption{Performance comparison in Top-k accuracy and MRR on LY Corporation datasets}
\label{Table4}
\resizebox{\linewidth}{!}{
\begin{tabular}{lcccc|cccc|cccc}
\hline
                 & \multicolumn{4}{c|}{Familiar Movement}          & \multicolumn{4}{c|}{Unfamiliar Movement}       & \multicolumn{4}{c}{Total Movement} \\ \cline{2-13} 
                 & Top-1 & Top-5 & Top-10 & MRR   & Top-1 & Top-5 & Top-10 & MRR   & Top-1 & Top-5 & Top-10 & MRR   \\\hline
Majority Vote          &    0.0477    &   0.1327     &     0.1903   &    0.0851    &        0.0120    &   0.0271     &     0.0395   &    0.0186        &   0.0299  &   0.0800 &   0.1151 &   0.0519   \\
MLP          &    0.0263    &    0.0933    &   0.1490     &    0.0561    &     0.0159   &    0.0492    &    0.0759    &    0.0308    &    0.0212    &    0.0717    &   0.1132     &    0.0437    \\
LSTM              &  0.0856      &    0.2504   &  0.3604      &    0.1570    &    0.0891    &    0.2025    &     0.2799   &    0.1397    &   0.0873     &     0.2269   &   0.3209     &   0.1485     \\
DeepMove              &  0.0933      &    0.2397    &  0.3437      &    0.1579    &  0.0808      &    0.1992    &  0.2737  &    0.1330    &   0.0872     &     0.2199   &   0.3095     &   0.1458     \\
STAN              &   \textbf{0.0956}      &    0.2630     &    0.3644     &   0.1668      &   0.0797     &   0.2188     &    0.2896    &    0.1384    &    0.0863     &   0.2407    & 0.3265 &   0.1522     \\
GETNext          &    0.0453    &  0.2007      &   0.3457     &    0.1148    &    0.0686    &    0.2070    &    0.3163    &    0.1283    &   0.0567     &    0.2038    &    0.3313    &    0.1214    \\
Transformer              &  0.0804      &    0.3014    &  0.4645      &    0.1770    &    0.0895    &    0.2827    &     0.4107   &    0.1737    &   0.0848     &     0.2922   &   0.4381     &   0.1754     \\
Informer             &    0.0690    &   0.3038     &  0.4768      &   0.1702     &   0.0725     &  0.2778      &    0.4183    &   0.1591     &    0.0707    &   0.2911     &   0.4481     &    0.1648    \\
MoE-TransMov        &   0.0807   &   \textbf{0.3238}     &      \textbf{0.4902}  &   \textbf{0.1826}     &    \textbf{0.1179}    &     \textbf{0.3250}   &  \textbf{0.4658} &   \textbf{0.2072}     &   \textbf{0.0989}     &   \textbf{0.3244}     &   \textbf{0.4782}     &    \textbf{0.1946 }   \\
\hline
\end{tabular}%
}
\end{table*}

As presented in Table \ref{Table4}, the proposed MoE-TransMov model consistently outperforms baseline approaches across most metrics on the LY dataset. For the Familiar Movement subset, STAN achieves the highest Top-1 accuracy (0.0956), highlighting its effectiveness in scenarios where historical user patterns are abundant and mobility behaviors are relatively regular. Nevertheless, MoE-TransMov surpasses all models in Top-5 (0.3238), Top-10 (0.4902), and MRR (0.1826), indicating superior capability in retrieving and ranking a broader set of relevant POIs beyond the most frequent choices. Performance on the Unfamiliar Movement subset, which poses a greater challenge due to limited historical interactions and increased behavioral variability, shows a noticeable decline across all baseline models. In contrast, MoE-TransMov exhibits remarkable robustness, achieving the best results in Top-1 (0.1179), Top-5 (0.3250), Top-10 (0.4658), and MRR (0.2072). These results demonstrate the model’s strong generalization ability under sparse and heterogeneous mobility conditions, benefiting from expert-based routing to capture diverse movement patterns. On the Total Movement dataset, MoE-TransMov again demonstrates its superiority, attaining the highest scores in Top-5 (0.3244), Top-10 (0.4782), and MRR (0.1946), confirming the effectiveness of the MoE framework in jointly modeling both familiar and unfamiliar movement behaviors.

Overall, these results substantiate the advantage of integrating Transformer-based shared encoders with a Mixture-of-Experts architecture for next-POI prediction in realistic urban environments. Although MoE-TransMov is trained solely on the total movement dataset, it consistently achieves state-of-the-art performance across familiar, unfamiliar, and total movement subsets, demonstrating its capacity to disentangle heterogeneous mobility patterns within a unified framework. Moreover, when compared with the results on the Foursquare NYC dataset, the performance gains of MoE-TransMov are more pronounced on the LY dataset, which is larger in scale, richer in POI diversity, and characterized by longer observation periods and less extreme popularity concentration. Such real-world large-scale mobility data inherently exhibit greater variability and complexity, making expert specialization particularly beneficial. The consistently strong performance of MoE-TransMov across different movement regimes on the LY dataset confirms its robustness, generalization capability, and practical applicability to large-scale, real-world next-POI prediction tasks.

\begin{figure}[h]
    \centering
    \includegraphics[width=1\textwidth]{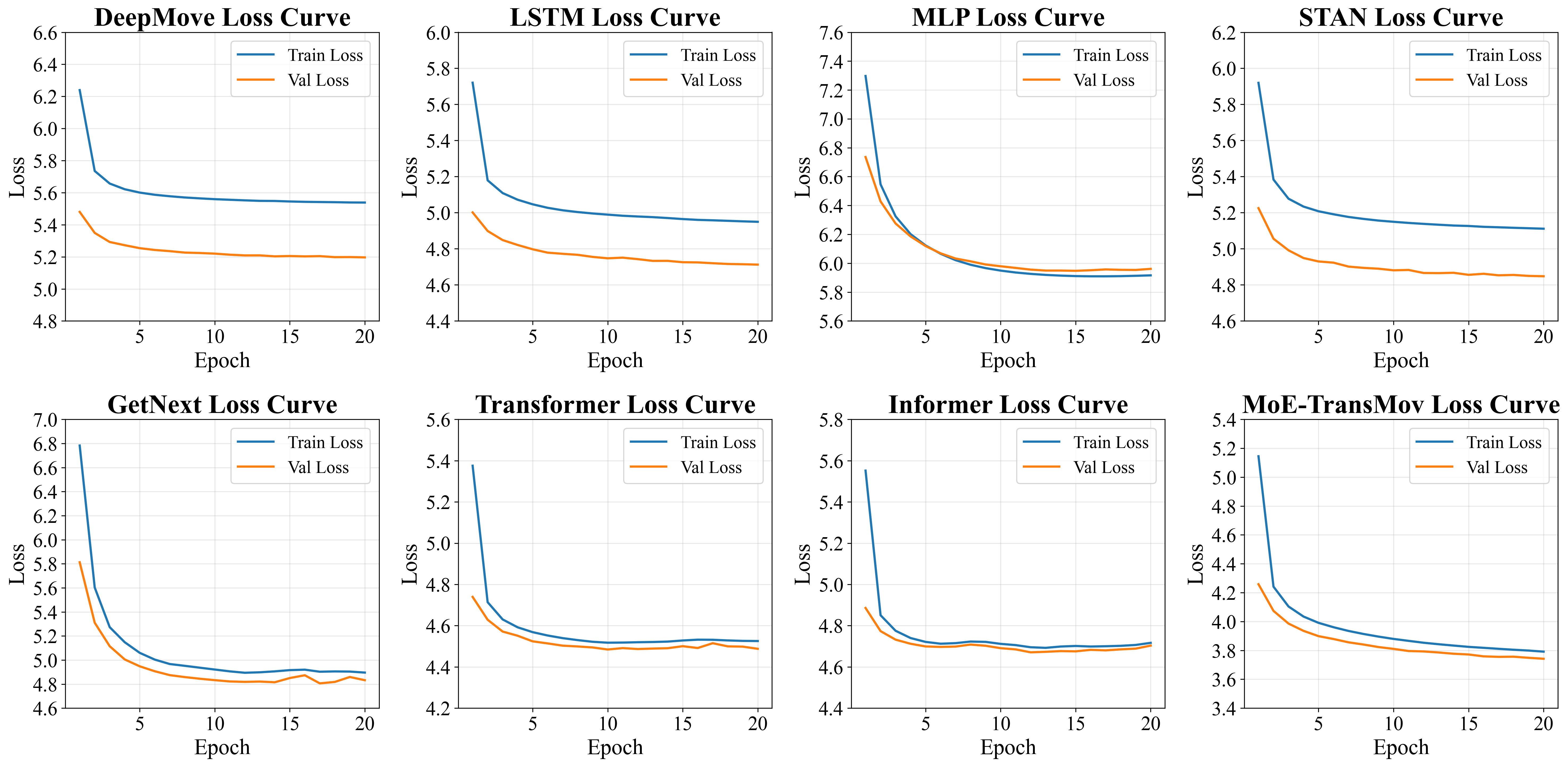} 
    \caption{Training and validation loss convergence of different Next POI prediction models on the Kyoto Mobility Dataset.} 
    \label{fig:figure4} 
\end{figure}

\subsection{Ablation Study}

\begin{table}[h]
\centering
\caption{Ablation Study for MoE-TransMov Model}
\label{Table5}
\resizebox{\linewidth}{!}{
\begin{tabular}{lcccc|cccc|cccc}
\hline
                 & \multicolumn{4}{c|}{Familiar Movement}          & \multicolumn{4}{c|}{Unfamiliar Movement}       & \multicolumn{4}{c}{Total Movement} \\ \cline{2-13} 
                 & Top-1 & Top-5 & Top-10 & MRR   & Top-1 & Top-5 & Top-10 & MRR   & Top-1 & Top-5 & Top-10 & MRR   \\\hline
w/o MoE              &  0.0804      &    0.3014    &  0.4645      &    0.1770    &    0.0895    &    0.2827    &     0.4107   &    0.1737    &   0.0848     &     0.2922   &   0.4381     &   0.1754     \\
w/o Transformer        &    0.0553    &   0.2680     &      0.4498  &   0.1490     &    0.0853    &     0.2885   &  0.4304 &   0.1748     &   0.0700     &   0.2781     &   0.4403     &    0.1617    \\
MoE(2 LSTMs)-TransMov        &   0.0794   &   0.2044     &      0.4778  &   0.1805     &    0.1093    &     0.3184   &    0.4586     &   0.2005    &   0.0940     &   0.3113     &    0.4584  & 0.1903  \\
MoE(2 Transformers)-TransMov        &   \textbf{0.0847}   &   0.3134     &      0.4875  &   \textbf{0.1846}     &    0.1013    &     0.3039   &  0.4481 &   0.1913     &   0.0928    &   0.3087     &   0.4682     &    0.1879    \\
MoE(1 Transformer, 1 LSTM)-TransMov        &   0.0807  &   \textbf{0.3238}     &      \textbf{0.4902}  &   0.1826     &    \textbf{0.1179}   &     \textbf{0.3250}   &  \textbf{0.4658} &   \textbf{0.2072}     &   \textbf{0.0989}     &   \textbf{0.3244}     &   \textbf{0.4782}     &    \textbf{0.1946 }   \\
\hline

\end{tabular}%
}
\end{table}

To evaluate the contribution of each architectural component in the proposed model, we conduct an ablation study with four variants. The first variant removes the MoE module, retaining only the shared Transformer layers. This setup isolates the effect of expert specialization on modeling heterogeneous movement patterns. The second variant omits the shared Transformer layers preceding the MoE module, thereby removing the global encoding that supports cross-context alignment. The third and fourth variants replace the expert modules with two LSTM-based or two Transformer-based experts, respectively, to examine the influence of the expert architecture on performance.

Table \ref{Table5} presents the results of our ablation study, which evaluates the impact of different components in the proposed MoE-TransMov model. The experiments systematically remove or modify key architectural elements to assess their individual contributions across familiar, unfamiliar, and total movement datasets. Removing the MoE component results in a performance decline across all metrics. Notably, for the unfamiliar movement dataset, Top-10 drops from 0.4658 (our proposed model) to 0.4107, and MRR decreases from 0.2072 to 0.1737. This demonstrates that the MoE module is particularly effective at handling less predictable, diverse user trajectories. Its ability to specialize through expert selection allows the model to better generalize to non-routine movements. When the Transformer is removed, the degradation is even more pronounced, especially in Top-1 accuracy, which falls from 0.0989 to 0.0700 on the total dataset. This underscores the critical role of self-attention in capturing both short- and long-range dependencies in user movement sequences. Without the Transformer, the model’s capacity to understand temporal relationships in mobility patterns is significantly impaired. We also compare variations of the MoE design: using two LSTM-based experts versus two Transformer-based experts or a hybrid of one Transformer and one LSTM. Among these, the hybrid architecture (MoE (1 Transformer, 1 LSTM)) achieves the best overall performance, especially on the unfamiliar movement dataset, which yields the highest Top-1 (0.1179), Top-5 (0.3250), Top-10 (0.4658), and MRR (0.2072). This suggests that combining long-range sequence modeling (Transformer) with short and sparse sequences modeling (LSTM) leads to stronger generalization. Overall, the full MoE-TransMov model with heterogeneous experts (Transformer + LSTM) outperforms all other variants, demonstrating that the Transformer and MoE components work synergistically. Their combination enables the model to robustly learn mobility patterns across different regions and movement contexts.

Unlike a hard conditional ensemble, our model is trained end-to-end with a shared encoder and soft expert combination, as evidenced by the superior performance over both two-LSTM and two-Transformer variants. These results validate that MoE enhances adaptability in different data scenarios, while Transformer-based attention improves ranking and sequence modeling, making our approach highly effective for next-POI prediction.

\section{CONCLUSION}

In this study, we proposed MoE-TransMov, a novel Transformer-based architecture enhanced with a MoE framework for next POI prediction. The model is specifically designed to differentiate between familiar and unfamiliar movement patterns, which is an essential yet often overlooked aspect of human mobility modeling. Our model addressed one of the key challenges in mobility modeling, that behavioral variability under different user contexts. By integrating a dynamic gating mechanism with self-attention, MoE-TransMov is able to adaptively route input sequences to the most suitable expert, capturing both short-term and long-term dependencies in user trajectories.

Comprehensive experiments conducted on two real-world check-in datasets: Foursquare NYC and a large-scale dataset of Kyoto, Japan provided by LY Corporation, which demonstrate the robustness and generalization ability of the proposed framework. On the Foursquare dataset, MoE-TransMov achieves state-of-the-art performance on both the Unfamiliar and Total Movement subsets, significantly outperforming all sequential and Transformer-based baselines. An interesting and unexpected observation is that the Majority Vote heuristic achieves the best Top-k accuracy on the Familiar subset, revealing that Foursquare users tend to repeatedly visit a small number of highly popular POIs. This popularity-driven pattern allows simple frequency-based predictions to be highly competitive. Nevertheless, MoE-TransMov still produces the strongest ranking quality (MRR) on this subset, indicating that it better captures fine-grained user preference ordering even in repetitive mobility contexts. In contrast, on the larger, richer, and more realistic LY dataset, which is characterized by more users, more POIs, and longer temporal spans, and the superiority of MoE-TransMov becomes even more evident. The model achieves the best performance across nearly all metrics and all three movement subsets, including Familiar, Unfamiliar, and Total Movement. These results confirm that the MoE architecture scales effectively to complex mobility environments and is highly capable of modeling diverse behavioral patterns that arise in real-world, multi-user, long-term datasets.

In the future, several promising research directions can further enhance the effectiveness and generalizability of MoE-TransMov. First, we plan to incorporate additional contextual modalities, such as weather conditions, temporal anomalies, such as holidays or emergencies, and user intent derived from web search data, to enable more context-aware decision making. Second, we aim to scale the number of expert modules and explore hierarchical gating mechanisms, which could facilitate a more granular understanding of latent mobility factors and regional behavior clusters. Furthermore, we intend to investigate the model's transferability by extending it to cross-city or cross-country POI prediction, examining how learned mobility patterns adapt across different urban structures and cultural contexts. Finally, integrating social signals and group behaviors, such as co-location or peer influence, presents an opportunity to enhance personalization and collaborative recommendation in mobility services. Through these extensions, we envision MoE-TransMov as a foundational framework for advancing robust, scalable, and context-aware human mobility prediction, with wide-ranging applications in intelligent transportation systems, urban planning, and location-based services.

\bibliographystyle{plainnat}
\bibliography{refe}
\end{doublespace}
\end{document}